\documentclass{article}
\usepackage{ijcai18}

\usepackage{times}
\usepackage{xcolor}
\usepackage{soul}
\usepackage[utf8]{inputenc}
\usepackage[small]{caption}
\usepackage{graphicx}
\usepackage{hyperref}
\usepackage{amsmath}
\usepackage{amssymb}
\usepackage{bm}
\usepackage{booktabs}
\usepackage{algorithm}  
\usepackage{algorithmic}
\usepackage{multirow}

\title{Visual Data Synthesis via GAN for Zero-Shot Video Classification}

\author{{Chenrui Zhang} and {Yuxin Peng}\thanks{Corresponding author.} \\ Institute of Computer Science and Technology, Peking University \\ Beijing 100871, China \\
pengyuxin@pku.edu.cn}

\begin{document}

\maketitle

\begin{abstract}
Zero-Shot Learning (ZSL) in video classification is a promising research direction, which aims to tackle the challenge from explosive growth of video categories. Most existing methods exploit seen-to-unseen correlation via learning a projection between visual and semantic spaces. However, such projection-based paradigms cannot fully utilize the discriminative information implied in data distribution, and commonly suffer from the information degradation issue caused by ``heterogeneity gap''. In this paper, we propose a visual data synthesis framework via GAN to address these problems. Specifically, both semantic knowledge and visual distribution are leveraged to synthesize video feature of unseen categories, and ZSL can be turned into typical supervised problem with the synthetic features.
First, we propose \textit{multi-level semantic inference} to boost video feature synthesis, which captures the discriminative information implied in joint visual-semantic distribution via feature-level and label-level semantic inference.
Second, we propose \textit{Matching-aware Mutual Information Correlation} to overcome information degradation issue, which captures seen-to-unseen correlation in matched and mismatched visual-semantic pairs by mutual information, providing the zero-shot synthesis procedure with robust guidance signals. Experimental results on four video datasets demonstrate that our approach can improve the zero-shot video classification performance significantly.
\end{abstract}

\section{Introduction}

In the past decade, supervised video classification methods have achieved significant progress due to deep learning techniques and large-scale labeled datasets. However, with the number of video categories growing rapidly, classic supervised frameworks suffer from the following challenges: (1) They rely on large-scale labeled data heavily, while collecting labels for video is laborious and costly, as well as the video instances of some categories are quite rare. (2) Particular models learned on limited categories are hard to expand automatically as video categories grow.

To overcome such restrictions, Zero-Shot Learning (ZSL) has been studied widely~\cite{Ziming:zhang2015zero,Soravit:changpinyo2016synthesized,Xun:xu2017transductive,Li:zhang2017learning}, which aims to construct
classifiers dynamically for novel categories. Instead of treating each category independently, ZSL adopts semantic representation (e.g., attribute or word vector) as side information, to associate categories in source domain and target domain for semantic knowledge transfer. Existing zero-shot classification methods focus primarily on \textit{still images}. In this paper, we focus on zero-shot classification in \textit{videos}, which is
more needed as novel video categories emerge on the web everyday. Furthermore, compared to \textit{image}, zero-shot \textit{video} classification has its own characteristics and thus is more challenging. First, video data contains more noise than image, setting a higher requirement on the robustness of zero-shot classification models. Second, video feature describes both spatial and temporal information, whose manifold is more complex. Third, video context with various poses and appearances can be changeable, leading to that video distribution is much more long-tailed than that of image.

Most existing ZSL methods learn a projection among different representation (i.e., visual, semantic or intermediate) spaces based on the data of seen domain, and the learned projection will be applied directly to unseen domain during testing. However, such \textit{projection-based} methods have several shortcomings and thus not robust enough for video ZSL. On the one hand, they exploit seen-to-unseen correlation only from the semantic knowledge aspect, while ignoring the discriminative information implied in visual data distribution. In fact, intrinsic visual distribution play a vital role in zero-shot
video classification, since the most discriminative information is derived from visual feature space~\cite{Long_2017_CVPR}. On the other hand, the inconsistency between visual feature and semantic representation causes \textit{heterogeneity gap}, and the projection between heterogeneous spaces leads to an information loss such that seen-to-unseen correlation would degrade. A specific affect caused by this gap is \textit{hubness problem}~\cite{radovanovic2010hubs}, where some irrelevant category prototypes become near neighbors of each other, as the projection is performed in high-dimensional spaces.

In this paper, we introduce Generative Adversarial Networks (GANs)~\cite{goodfellow2014generative} to
realize zero-shot video classification
from the perspective of generation, which bypasses the above limitations of explicit projection learning.
The key idea is to model the joint distribution over high-level video feature
and semantic knowledge via adversarial learning, where discriminative information and seen-to-unseen correlation are
embedded effectively for novel video feature synthesis. Once the training procedure is done, unseen video feature can be synthesized by the generator, and zero-shot classification is realized in video feature space in a supervised fashion. Namely,
synthetic video features are utilized to train a conventional supervised classifier (e.g., SVM), or serve directly to the simplest
nearest neighbor algorithm.

However, synthesizing discriminative video feature based
on semantic knowledge is non-trivial for prevalent adversarial
learning framework such as conditional GAN (cGAN)~\cite{mirza2014conditional}.
The main challenges we encountered are two-fold: (1) How to model the joint distribution over
video feature and semantic knowledge robustly and ensure the
discriminative characteristics of the synthetic feature? (2) How to mitigate the impact of heterogeneity gap and transfer
semantic knowledge maximally?

To tackle the first challenge, we propose \textit{multi-level semantic inference} approach, aiming to fully
utilized the distribution over video feature and semantic knowledge for discriminative information mining. It contains two opposite synthesis procedures
driven by adversarial learning, where the semantic-to-visual branch synthesizes video feature given semantic knowledge, and the visual-to-semantic branch inversely infers the semantic knowledge at both feature-level
and label-level. Such two-pronged semantic inference forces the generator to capture the discriminative attributes for visual-semantic alignment, and bidirectional synthesis procedures boost each other collaboratively
for ensuring robustness of the synthetic video
feature.

To tackle the second challenge, we propose \textit{matching-aware mutual information correlation}, which provides the synthesis procedure with informative guidance signals for overcoming the information degradation issue. Instead of direct feature projection, the mutual information hints among matched and mismatched visual-semantic pairs are utilized for semantic knowledge transfer. Therefore, statistical dependence among heterogeneous representations can be  captured, bypassing the information degradation issue in typical
projection-based ZSL methods.

To verify the effectiveness of the proposal, we conduct extensive experiments on 4 widely-used video datasets, and the experimental results demonstrate that our approach improves the zero-shot video classification performance significantly.

\section{Related Work}
\subsection{Zero-shot Learning}
Zero-shot learning has been drawn a wide attention, due to its potentiality on scaling to large novel categories in classification tasks. Early explorations in ZSL are mainly focus on learning probabilistic attribute classifiers (PAC), such as Direct-Attribute Prediction (DAP)~\cite{lampert2009learning}, Indirect Attribute Prediction (IAP)~\cite{lampert2014attribute} and CONSE~\cite{norouzi2014zeroshot}. In these approaches, posterior of seen categories is predicted firstly, then the attribute classifiers are learned by the principle of maximizing posterior estimation. PAC has been proved to be poor in ZSL tasks as it ignores the relation of different attributes~\cite{jayaraman2014decorrelating}. After that, research efforts shift to another direction which directly builds a projection from visual feature space to semantic (e.g. attribute or word-vector) space. In this paradigm, linear~\cite{palatucci2009zero}, bilinear~\cite{Bernardino:romera2015embarrassingly,zhang2016zero} and nonlinear~\cite{socher2013zero,Li:zhang2017learning} compatibility models are explored widely by optimizing specific loss functions. Besides visual $ \rightarrow $ semantic mapping, other two mapping directions are studied, namely semantic $ \rightarrow $ visual mapping and embedding both visual and semantic features into another shared space~\cite{Ziming:zhang2015zero}. Nowadays, various hybrid models~\cite{akata2016multi,Soravit:changpinyo2016synthesized} are also studied based on such diverse selection of embedding space.

Recently, Unseen Visual Data Synthesis (UVDS)~\cite{Long_2017_CVPR} views ZSL from a new perspective, i.e., it tries to synthesize visual feature of unseen instances for converting ZSL to a typical supervised problem. However, it is still an explicit projection learning framework in which visual feature is synthesized through embedding-based matrix mapping. On the one hand, such explicit projection paradigm is hard to preserve manifold information of visual feature
space, or capture unseen-to-seen correlation for zero-shot generalization. On the other hand, the visual feature synthesized by UVDS suffers variance decay issue~\cite{Long_2017_CVPR}, which is not robust enough for high-dimensional video feature synthesis. In this paper, we address these problems with deep generative models.

\subsection{Generative Adversarial Networks}
As one of the most potential generative models, Generative Adversarial Networks (GANs)~\cite{goodfellow2014generative} are studied extensively recently. The goal of GAN is to learn a generator distribution $ P_g(x) $ that matches the real data distribution $ P_{real}(x) $ through a minimax game:
\begin{multline*}
\setlength{\abovedisplayskip}{-10pt}
\min\limits_{G}\max\limits_{D}V(G,D)=\mathbb{E}_{x\sim P_{real}}[\log D(x)] \\
+\mathbb{E}_{z\sim P_{noise}}[\log {(1-D(G(z)))}]
\vspace*{-80pt}
\end{multline*}
\begin{figure*}[htb]
	\begin{center}
		\includegraphics[height=5.2cm]{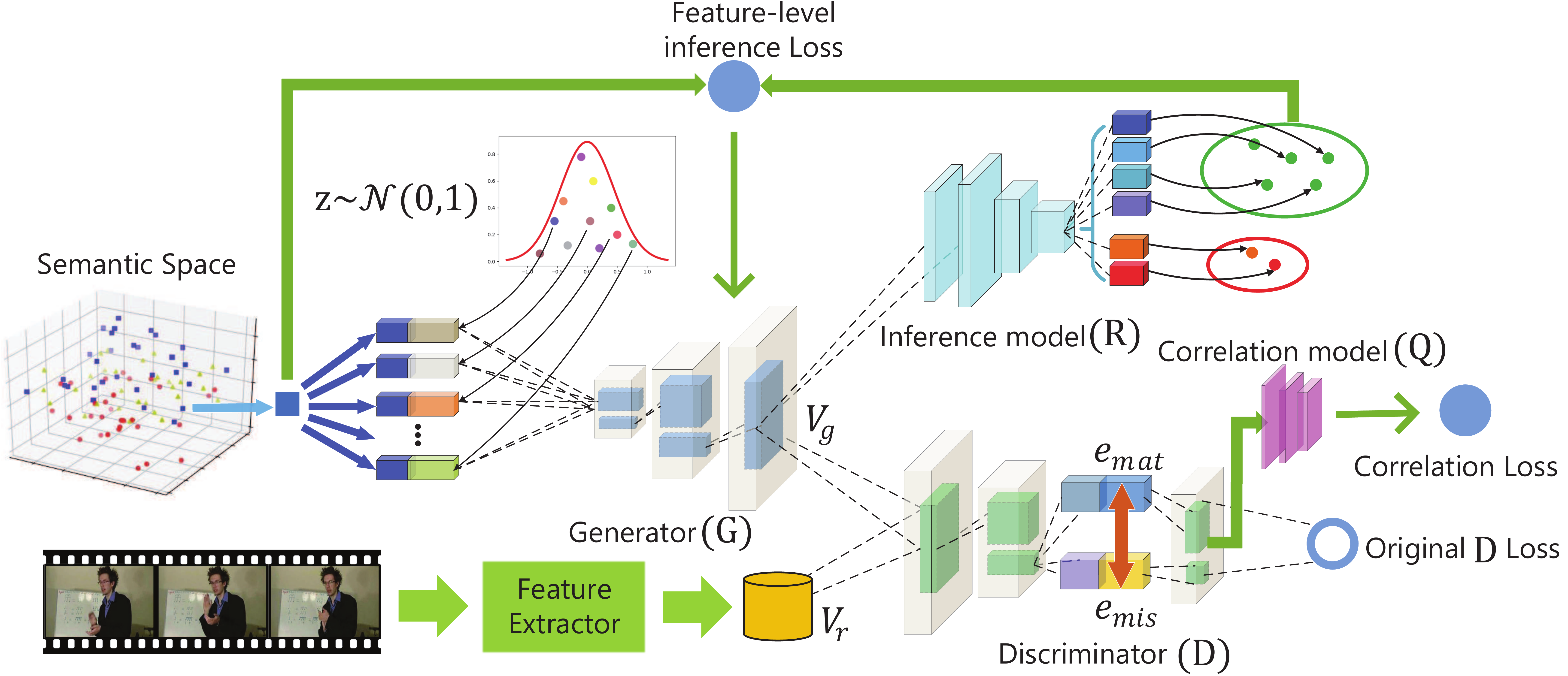}
	\end{center}
	\vspace{-1.4em}
	\caption{Architecture of the proposed framework. A group of noise is utilized by generator to synthesize video feature $ V_g $, which is used by inference model $R$ and discriminator $D$ simultaneously to perform semantic inference and correlation constraint. $ V_r $ denotes the real visual feature, $ e_{mat} $ and $ e_{mis} $ respectively denote the matched and mismatched semantic embedding sets corresponding to $ V_r $.}
	\label{fig:short}\vspace{-1.0em}
\end{figure*}

Basic GAN is unrestricted and uncontrollable during training, as well as the priori noise $ z $ is difficult to explain. To tackle these defects, a surge of variants of GAN have emerged, such as conditional GAN (cGAN)~\cite{mirza2014conditional}, InfoGAN~\cite{chen2016infogan} and Wasserstein GAN (W-GAN)~\cite{arjovsky2017wasserstein}. cGAN tries to solve the controllability issue via providing class labels to both generator and discriminator as condition. InfoGAN aims to learn more interpretable representation by maximizing the mutual information between latent code and generator's output. W-GAN introduces earth mover distance to measure similarity of real and fake distribution, which makes training stage more stable. Our work is related to cGAN and InfoGAN closely, and details are discussed in the sequel.

\section{Methodology}

\subsection{Problem Formalization}

Assume $D_{tr}=\{\left(V_n,y_n\right)\}_{n=1}^{N_s}$ denotes the training set of $ N_s $ samples and $ D_{te}=\{\left(V_n,y_n\right)\}_{n=1}^{N_u} $ denotes test set of $ N_u $ samples, their corresponding label spaces are $ \mathcal{S}=\{1,2,\cdots,S\} $ and $ \mathcal{U}=\{S+1,S+2,\cdots,S+U\} $ with $\mathcal{S}\cap\mathcal{U}=\varnothing$ in ZSL context. \(V_n\) and \(y_n\) respectively denote visual feature and label of the \(n^{st}\) video sample. Given a new test video feature $ V $, the goal of ZSL is to predict its class label $ y \in \mathcal{U} $. We use \(g(\cdot):Y\rightarrow E\) to denote the word embedding function that maps label to semantic embedding and $ g(y_i) $ is the word-vector corresponding to label $ y_i $. $ Y $ and $ E $ denote label space and semantic space respectively.

\subsection{Multi-level Semantic Inference}
As illustrated in Fig. \ref{fig:short},
the generator $G$ synthesizes video feature conditioned on the semantic knowledge (i.e., word vector in this paper). The
synthetic video feature is expected to be convincing enough to recover the distribution of real video data, as well as capture
the semantic correlation and discriminative information for zero-shot classification task. Essentially, the mission of $ G $ is learning a \textit{visual-semantic joint distribution}, rather than fitting a single distribution of only one domain. However, high-dimensional
and noisy nature of video feature lead to great
uncertainty for visual-semantic matching, and thus conventional GAN frameworks are hard to achieve stable synthesis. Moreover, stable visual-semantic matching cannot ensure the discriminative power of the synthetic feature, which is vital for zero-shot classification task. 

To tackle the above challenges, we propose a multi-level semantic inference approach, which boosts the video feature synthesis via inversely inferring semantic information at both representation-level and label-level. Video feature synthesis and semantic inference are driven by adversarial learning, where semantic inference forces the generator to capture the seen-to-unseen correlation and discriminative information, boosting the video feature synthesis for more robust visual-semantic alignment.

Concretely, we develop an auxiliary model $R$ for semantic feature inference, which learns an inverse mapping from the synthetic video feature to corresponding semantic knowledge. Formally, $ R $ tries to fit conditional distribution $ q(e_c|\hat{v}) $, where $ \hat{v} $ is the synthesized video feature and $ e_{c} $ is the semantic . In fact, $ G $ and $ R $ model two joint distributions:

\begin{equation}
\setlength{\abovedisplayskip}{-2pt}
\setlength{\belowdisplayskip}{-2pt}
\left\{
\begin{array}{lr}  
p_{\theta}(v,z,e_c)=p_{\theta}\,(z,e_c)\,p_{\theta}(v|z\;\!,e_c)   \\
q_{\phi}(v,z,e_c)=q_{\phi}(z,v)\,q_{\phi}(e_c\;\!|z,v) 
\end{array}
\right.
\label{eq:two_joint_dist}
\end{equation}where $ \theta $ and $ \phi $ are parameters of $ G $ and $ R $, respectively. As semantic and video features are high-level representation rather than real-valued data (e.g., image pixels), typical reconstruction metrics such as $\ell_p$-norm is hard to measure the semantic similarity. Hence, we adopt adversarial learning for feature-level semantic inference:
\begin{equation}
\begin{aligned}
\mathcal{F}_{fea}&=\min\limits_{\theta,\phi}\max\limits_{\psi}V(\theta,\phi,\psi) \\
&=\mathbb{E}_{v\sim{q_{\phi}(v)},\hat{c}\sim{q_{\phi}(c|\hat{v})}}\big[\log{D_{\psi}(v,\hat{c})}\big] \\
&+\mathbb{E}_{\hat{v}\sim{p_{\theta}(v|z,c)},(z,c)\sim{p_{\theta}(z,c)}}\big[\log{\big(1-D_{\psi}(\hat{v},c)\big)}\big]
\end{aligned}
\label{eq:ALI_like}
\end{equation}
where $\mathcal{F}_{fea}$ is the adversarial object of feature inference, $ \psi $ is the parameter of the discriminator $ D $, the meaning of $ \theta $ and $ \phi $ is same with Eq.~\eqref{eq:two_joint_dist}.

For discriminative information mining, we enhance minimax game of Eq.~\eqref{eq:ALI_like} with category log-likelihood estimation, which can be viewed as \textit{label-level} semantic inference:
\begin{equation}
\setlength{\abovedisplayskip}{-0.7pt}
\setlength{\belowdisplayskip}{-0.7pt}
\mathcal{F}_{cat}\!=\!\sum\limits_{i=1}\limits^{N_s}\mathbb{E}\big[\!\log{P\big(C\!=C_i|v\big)}\big]\!+\!\mathbb{E}\big[\log{P\big(C=C_i|\hat{v}\big)}\!\big]
\label{eq:class_adversarial}
\end{equation}
where $\mathcal{F}_{cat}$ is the objective of category-aware training, $ C_i $ denotes the $ i $-th category, $ N_s $ is the total number of seen categories and $ P $ means the category-specific distribution. The insights behind Eq.~\eqref{eq:class_adversarial} is that label judgment can provide valuable information for decision boundary selection of category-level distribution, guiding $ G $ to synthesis visual features with more discriminative information.

In addition, to further stabilize such bidirectional adversarial synthesis, we design a simple yet effective outlier detection approach. As shown in Figure~\ref{fig:short}, rather than sampling only one noise $ z $ for each condition $ c $, a group of noises are sampled randomly to generate several video features $ \mathcal{V}_g $. $ R $ infers semantic features $ \mathcal{R} $ with $ \mathcal{V}_g $, then the similarity among elements in $ \mathcal{R} $ is measured via cosine distance, outliers (denoted by $ \mathcal{O} $) that exceed an adaptive threshold $ \eta $ will be discarded. We call $ \mathcal{P}=\mathcal{R}-\mathcal{O} $ is the set of \textit{reasonable} inferred word vectors, and for arbitrary $ r_i\in \mathcal{R} $, $ r_i $ is reasonable if and only if:
\begin{equation}
\setlength{\abovedisplayskip}{-0.7pt}
\setlength{\belowdisplayskip}{-0.7pt}
Cos(e_{c},r_i)\ge\eta=\max\limits_{1\le j\le n}\big(Cos(e_{c},r_j)\big)-\mu\Delta
\end{equation}
where $\mu$ is trade-off parameter and $ \Delta $ is the variation range of cosine similarity among elements in $ \mathcal{R} $ during training:
\begin{equation}
\setlength{\abovedisplayskip}{-0.3pt}
\setlength{\belowdisplayskip}{-0.3pt}
\Delta=\max\limits_{1\le j\le n}Cos(e_{c},r_j)-\min\limits_{1\le j\le n}Cos(e_{c},r_j)
\end{equation}
Thus, $ \eta $ adjusts adaptively along with the training progress to boost the robustness of semantic inference.
Finally, We define the inference loss as:
\begin{equation}
\setlength{\abovedisplayskip}{-0.7pt}
\setlength{\belowdisplayskip}{-0.7pt}
\mathcal{L}_{re}(e_{c},\mathcal{P})=-\sum_{r_i\in \mathcal{P}}Cos(e_{c},r_i)
\label{equ:rec0}
\end{equation}

\subsection{Mutual Information Correlation Constraint}

The power of GAN is attributed to that it learns a latent loss by $ D $ to classify real or fake instances in a data-driven manner, rather than relying on structured loss which is hand-engineered. This powerful latent loss guides $ G $ to fit sophisticated distribution of real data progressively. However, in ZSL video classification scenarios, $ D $ should learn to evaluate whether the output of $ G $ is aligned with conditional information, instead of only scoring its realism. Original loss of cGAN is not informative enough for $ G $ to synthesis discriminative visual feature data since $ D $ has no explicit notion of whether visual feature matches the semantic embedding. To address this problem, we propose Matching-aware Mutual Information Correlation Constraint (MMICC) to maximize semantic knowledge transfer through leveraging the Mutual Information (MI) of matched and mismatched visual-semantic pairs. Formally, MI can be defined based on entropy as:
\begin{equation}
I(X;Y)=H(X)-H(X|Y)=H(Y)-H(Y|X)
\label{eq:MI}
\end{equation}where $ X $ and $ Y $ are random variables, $ I(X;Y) $ denotes mutual information of $ X $ and $ Y $, $ H(X|Y) $ is conditional entropy:
\begin{equation}
\setlength{\abovedisplayskip}{0.1pt}
\setlength{\belowdisplayskip}{0.2pt}
H(X|Y)=-\sum\limits_{X\in \mathcal{X},Y\in\mathcal{Y}}\!{P(X,Y)\log P(X|Y)}
\label{eq:conditional_entropy}
\end{equation}
Eq. \eqref{eq:MI} intuitively reveals the meaning of MI, which quantitatively measures the amount of information given by one random variable about the other. 

We realize MMICC via a mutual information regularization defined as follows:
\begin{equation}
\mathcal{L}_{co}=-I\big(e_{mat};G(z,e_{mat})\big)+I\big(e_{mis};G(z,e_{mat})\big)
\label{eq:loss_co}
\end{equation}
where $ e_{mat} $ and $ e_{mis} $ respectively denote the matched and mismatched semantic embedding corresponding to specific visual feature. $ G(z,e_{mat}) $/$ G(z,e_{mis}) $ is the synthetic visual feature conditioned on $ e_{mat} $/$ e_{mis} $. Minimizing Eq. \eqref{eq:loss_co} represents that MI among matched pairs is expect to be high while among mismatched pairs is expect to be low, and this smooth matching-aware regularization provides $ G $ with guidance information during training.
\renewcommand{\algorithmicrequire}{\textbf{Input:}}
\renewcommand{\algorithmicensure}{\textbf{Output:}}
\begin{algorithm}[!h]
	\caption{Training process of the proposed framework}
	\label{alg:Framework}
	\begin{algorithmic}[1]
		\REQUIRE 
		minibatch visual feature $ v $, matched semantic embedding $ e $, the set of mismatched embedding
		$ \hat{E} $, the number of noise samples $ q $, training steps $ M $, step size $ m $.
		\FOR{$i=1$ \textbf{to} $M$}
		\STATE $ Score_r \leftarrow D(v,e) $ \{$\langle $real $ v $, matched $ e $$ \rangle $\}
		\STATE $ Score_w \leftarrow 0 $, $ \mathcal{R} \leftarrow \varnothing $ \{Initialization\}
		\FOR{\textbf{each} $ \hat{e_i} \in \hat{E} $}
		\STATE $ Score_w\!\!=\!\!Score_w\!+\!D(\!v,\!\hat{e_i}\!)$\,\{$ \!\langle $real\,$ v $,mismatched $ e $$ \rangle \!$\}
		\ENDFOR
		\STATE $ Score_w=Score_w/\#\hat{E} $ \{Average the $ Score_w $\}
		\FOR{$ i=1 $ \textbf{to} $ q $}
		\STATE $ z_i \sim \mathcal{N}(0,1)^{Z} $ \{Draw noise samples randomly\}
		\STATE $ \hat{v_i} \leftarrow G(z_i,e) $ \{Forward through generator\}
		\STATE $ Score_f\!=\!Score_f\!+\!D(\hat{v_i},\!e) $ \{$ \langle $fake $ v $, matched $ e $$ \rangle $\}\STATE $ r_i\! \leftarrow\!R(\hat{v_i}),  \mathcal{F}_{cat}\!\leftarrow\!D(v_i,\hat{v}_i) $ \!\{Semantic inference\}
		\STATE $ \mathcal{R}=\mathcal{R} \cup \{r_i\} $
		\ENDFOR
		\STATE $ Score_f = Score_f / q $ \{Average the $ Score_f $\}
		\STATE $ \mathcal{L}_{in}\!\leftarrow\! $ \eqref{equ:rec0}\,\{Get semantic inference loss via Eq. \eqref{equ:rec0}\}
		\STATE $ \mathcal{L}_{co}\!\leftarrow\! $ \eqref{eq:loss_co}\,\{Get correlation loss via Eq. \eqref{eq:loss_co}\}
		\STATE $ \mathcal{L_D} \leftarrow log(Score_r) + [log(1-Score_w)+log(1-Score_f)] / 2 + \lambda_2\mathcal{L}_{co} $
		\STATE $ D \leftarrow D-m\dfrac{\partial{\mathcal{L_D}}}{\partial{D}} $ \{Update discriminator\}
		\STATE $ \mathcal{L_G} \leftarrow log(Score_f) + \lambda_1\mathcal{L}_{in} $
		\STATE $ G \leftarrow G-m\dfrac{\partial{\mathcal{L_G}}}{\partial{G}} $ \{Update generator\}
		\ENDFOR
	\end{algorithmic}
\end{algorithm}

\newcounter{mytempeqncnt}
\begin{figure*}[!t]
	\normalsize
	\begin{equation}
	\begin{aligned}
	\min\limits_{G,Q}\max\limits_{D}V(G,D,Q)=&\mathbb{E}_{v,e\sim P(v,e)}\big[\log D(v|e_{mat})+\log (1-D(v|e_{mis}))\big]+ \\
	&\mathbb{E}_{z\sim P_{noise}}\big[\log {\big(1-D(G(z|e_{mat}))\big)}\big]+
	\lambda_1\mathcal{L}_{in}(e_{mat},G(v|e_{mat}))-\lambda_2\mathcal{L}_{co}(G,Q)
	\end{aligned}
	\label{eq:finalEq}
	\end{equation}
	\vspace*{-15pt}
\end{figure*}
With conditional entropy definition in Eq. \eqref{eq:conditional_entropy}, Eq. \eqref{eq:loss_co} can be further expressed as follows (we denote $ e_{mat} / e_{mis} $ as $ e $):
\begin{equation}
I\big(e;G(z,e)\big)\!=\!H(e)\!+\!{P(e,G(z,e))\log P(e|G(z,e))}
\end{equation}

In practice, the posterior $ P\big(e|G(z,e)\big) $ is hard to solve, thus $ I\big(e;G(z,e)\big) $ cannot be maximized directly. Fortunately the prior work~\cite{chen2016infogan} has solved this problem via \textit{Variational Information Maximization} and subsequent approximation. Specifically, they introduce an auxiliary distribution $ Q(c|x) $ to approximate original $ P(c|x) $ and defined a variational lower bound of MI based on $ Q(c|x) $. According to~\cite{chen2016infogan}, we use lower bound of MI, $ LB(G,Q) $, to approximate $ I\big(e;G(z,e)\big) $:
\begin{multline}
\setlength{\abovedisplayskip}{-0.7pt}
\setlength{\belowdisplayskip}{-0.7pt}
\begin{aligned}
LB(G,Q)&=\mathbb{E}_{e\sim P(e),x\sim G(z,e)}[\log Q(e|x)]+H(e) \\
&\le I\big(e;G(z,e)\big)
\end{aligned}
\end{multline}where $ Q $ is the auxiliary distribution which approximates original $ P\big(e|G(z,e)\big) $. $ H(e) $ is irrelevant to parameters of $ G $, thus we regard it as a constant during optimization. In our case, auxiliary distribution $ Q $ is represented by a neural network which shares all convolutional layers with $ D $ and yields conditional distribution $ Q(e|x) $ by additional three {\tt fc} layers.

\subsection{Objective and Optimization}
Comprehensively, objective of the proposed framework is summarized as Eq. \eqref{eq:finalEq}, where $ v $ is the visual feature, $ \lambda_1 $ and $ \lambda_2 $ are hyper-parameters which balance semantic inference loss $\mathcal{L}_{in}$ and correlation loss $\mathcal{L}_{co}$. We summarize the training procedure of our framework in Algorithm~\ref{alg:Framework}. Similar to~\cite{reed2016generative}, input of $ D $ is three-fold: 1) $ \langle v_{real},e_{mat} \rangle $ pair, 2) $ \langle v_{real},e_{mis} \rangle $ pair and 3) $ \langle v_{fake},e_{mat} \rangle $ pair. We use $ Score_r $, $ Score_w $  and $ Score_f $ to denote the confidence scores given by $ D $ corresponding to these three kind of pairs, respectively.

\subsection{Zero-shot Classification}
At the test stage, we simply use the nearest neighbor (NN) search and SVM as classifiers for evaluating the discriminative capability of synthetic video feature. For NN search, label in test set is predicted by:
\begin{equation}
\hat{y}=arg \min\limits_{y\in y_{te},v^*\in V_{te}}||G\big(z,g(y)\big)-v^*||
\end{equation}
where the $ G\big(z,g(y)\big) $ is the synthetic visual feature of label $ y $, $ v^* $ is the original visual feature in test set $ V_{te} $ and $ \hat{y} $ is the predicted label.

For experiments with SVM, we synthesize visual feature of same amount with unseen categories in test set. Then the synthetic feature of unseen categories and original visual feature of seen categories are merged to train a SVM with $ 3^{rd} $-degree polynomial kernel, whose slack parameter is set to 5.

\section{Experiments}
\subsection{Datasets and Settings}

\noindent\textbf{Datasets:} Experiments are conducted on four popular video classification datasets, including HMDB51~\cite{kuehne2013hmdb51}, UCF101~\cite{soomro2012ucf101}, Olympic Sports~\cite{niebles2010modeling} and Columbia Consumer Video (CCV)~\cite{jiang2011consumer}, which respectively contain 6.7k, 13k, 783 and 9.3k videos with 51, 101, 16 and 20 categories.
	
	\noindent\textbf{Zero-Shot Settings:} There are two ZSL settings, namely strict setting and generalized setting~\cite{xian2017zero}. The former assumes the absence of seen classes at test state while the latter takes both seen and unseen data as test data during testing. In this paper, we adopt strict setting for both NN search and SVM experiments.

\noindent\textbf{Data Split:} There are quite few zero-shot learning evaluation protocols for video classification in the community. Xu et al.~\cite{Xun:xu2017transductive} established a baseline of this filed. In order to compare our framework with the state-of-the-art, we follow the data splits proposed by~\cite{Xun:xu2017transductive}: 50/50 for every dataset, i.e., visual feature of 50\% categories are used for model training and the other 50\% categories are held unseen until test time. We take the average accuracy and standard deviation as evaluation metrics and report the results over 50 independent splits generated randomly.

\subsection{Implementation Details}
Our model is implemented with PyTorch\footnote{\tt http://pytorch.org/}. Both traditional feature and deep feature are investigated as video feature. For the former, similar to ~\cite{Xun:xu2017transductive}, we extract improved trajectory feature (ITF) with three descriptors (HOG, HOF and MBH) for each video, then encode them by Fisher Vectors (FV) and we get combined visual feature with 50688 dimension. For the latter, we use the two-stream~\cite{simonyan2014two} framework based on VGG-19 to extract spatial-temporal feature of videos, and frame and optical flow feature from last pooling layer are concatenated to form final visual embedding. We adopt GloVe~\cite{pennington2014glove} that trained on Wikipedia with more than 2.2 million unique vocabularies to obtain semantic embedding and its dimension is 300. The dimension of Gaussian noise is 100 and the cardinality of noise set is set to 30. We train our framework for 300 epochs using the Adam optimizer with momentum 0.9. We initialize the learning rate to 0.01 and decay it every 50 epochs by a factor of 0.5. Both $ \lambda_1 $ and $ \lambda_2 $ are set to 1.

\subsection{Compared Methods}
We compare our framework with several ZSL methods in video: (1) Convex Combination of Semantic Embeddings (CONSE)~\cite{norouzi2014zeroshot}. CONSE is a posterior based model which trains classifiers on seen categories, then the prediction models are built on linear combination of existing posterior. (2) Structured Joint Embedding (SJE)~\cite{akata2015evaluation}. SJE optimizes the structural SVM loss to learn the bilinear compatibility. This model utilized bilinear ranking to maximize the score among relevant labels and minimize the score among irrelevant labels. (3) Manifold regularized ridge regression (MR)~\cite{Xun:xu2017transductive}. MR enhances the convensional projection pipeline by manifold regularization, self-training and data augmentation in transductive manner.
\begin{table*}[htbp]
	\centering
	\setlength{\abovecaptionskip}{6pt}%
	\caption{Experimental results of our approach and comparison to state-of-the-art for ZSL video classification on four datasets. Average \% accuracy $ \pm $ standard deviation for HMDB51 and UCF101, mean average precision $ \pm $ standard deviation for Olympic Sports and CCV. All experiments use same word-vector as semantic embedding for fair consideration. Random guess builds the lower bound for each dataset. FV and DF denote the traditional and deep video feature respectively.}
	\label{tab:1}
		\begin{tabular}{lccccc}
			\toprule
			Methods & Feature & HMDB51 & UCF101 & Olympic Sports & CCV \\
			\midrule
			Random Guess & -- & 4.0 & 2.0 & 12.5 & 10.0\\
			\midrule
			CONSE~\cite{norouzi2014zeroshot} & FV & 15.0 $ \pm $ 2.7 & 11.6 $ \pm $ 2.1 & 36.6 $ \pm $ 9.0 & 20.7 $ \pm $ 3.1 \\
			SJE~\cite{akata2015evaluation} & FV & 12.0 $ \pm $ 2.6 & 9.3 $ \pm $ 1.7 & 34.6 $ \pm $ 7.6 & 16.3 $ \pm $ 3.1\\
			MR~\cite{Xun:xu2017transductive} & FV & 24.1 $ \pm $ 3.8 & 22.1 $ \pm $ 2.5 & 43.2 $ \pm $ 8.3 & 33.0 $ \pm $ 4.8 \\
			Ours (NN) & FV & 22.8 $ \pm $ 4.0 & 23.7 $ \pm $ 4.5 & 39.5 $ \pm $ 9.2 &  28.3 $ \pm $ 5.7\\
			Ours (SVM) & FV & \textbf{25.3 $ \pm $ 4.5} & \textbf{25.4 $ \pm $ 3.1} & \textbf{43.9 $ \pm $ 7.9} & \textbf{33.1 $ \pm $ 5.8}\\
			\midrule
			CONSE~\cite{norouzi2014zeroshot} & DF & 12.9 $ \pm $ 3.1 & 8.3 $ \pm $ 3.0 & 21.1 $ \pm $ 8.4 & 17.2 $ \pm $ 5.3 \\
			SJE~\cite{akata2015evaluation} & DF & 10.7 $ \pm $ 3.5 & 9.9 $ \pm $ 2.6 & 25.8 $ \pm $ 8.3 & 14.5 $ \pm $ 4.0 \\
			MR~\cite{Xun:xu2017transductive} & DF & 19.6 $ \pm $ 4.9  & 24.1 $ \pm $ 4.2 & 33.5 $ \pm $ 9.2 &  22.6 $ \pm $ 6.4\\
			Ours (NN) & DF & 18.5 $ \pm $ 5.3 & 27.3 $ \pm $ 5.9 & 30.1 $ \pm $ 9.5 & 22.3 $ \pm $ 7.6 \\
			Ours (SVM) & DF & \textbf{21.6 $ \pm $ 5.5} & \textbf{28.8 $ \pm $ 5.7} & \textbf{35.5 $ \pm $ 8.9} & \textbf{26.1 $ \pm $ 8.3} \\
			\bottomrule
		\end{tabular}
	\vspace{-0.6em}
\end{table*}
\begin{figure*}[htb]
	\begin{center}
		\includegraphics[height=3.1cm]{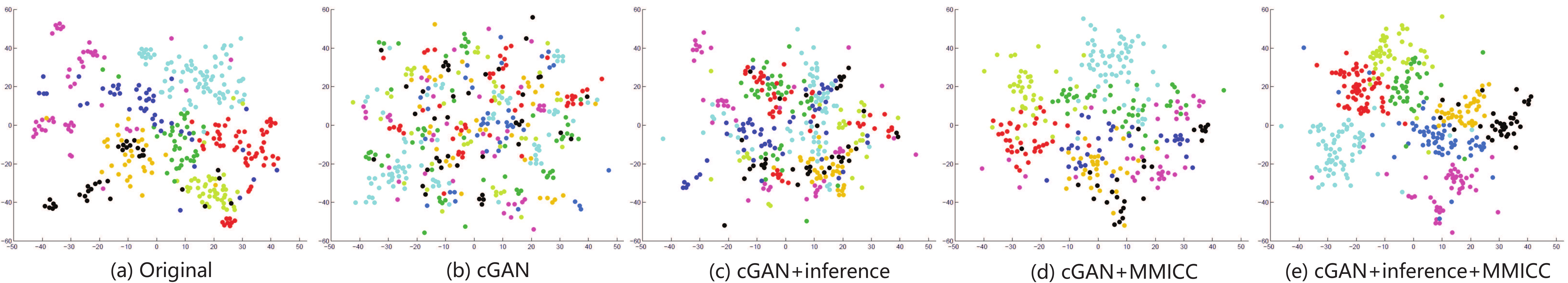}
	\end{center}
	\vspace{-1.0em}
	\caption{Visualization of the traditional visual feature distribution of 8 random unseen classes on Olympic Sports dataset. Different classes are shown in different colors.}\label{fig:visualize}
	\vspace{-0.9em}
\end{figure*}
\subsection{Experimental Results}
The experimental results are shown in Table~\ref{tab:1}. Note that FV denotes the Fisher Vectors encoded dense trajectory feature and DF denotes the deep video feature extracted by two-stream neural networks. From the results we draw several conclusions: (1) All methods are far beyond the random guess bound, demonstrating the success of ZSL in video classification. (2) Our approach beats the most existing methods in terms of average accuracy/mAP, while the standard deviation is slightly higher. This fact illustrates that generative framework is not as stable as projection-based methods, such as SJE and MR. (3) Compared to SVM, NN search suffers from more hubness problem and thus achieves suboptimal results. However, NN that based on our approach performs better than CONSE and SJE without extra pay, indicating the discriminative power of synthetic feature generated by our approach. (4) Our approach can improve the zero-shot video classification performance significantly both on traditional visual feature and deep video feature, indicating that our proposal is robust enough to capture the main discriminative information of different visual feature.

\subsection{Ablation Studies}
\begin{table}[htb]
	\centering
	\setlength{\abovecaptionskip}{2pt}%
	\setlength{\belowcaptionskip}{1pt}%
	\caption{Baseline experiments on performance of semantic inference and MMICC, which are denoted by Inf and Co respectively.}
	\label{tab:baselineTab}
	\begin{tabular}{llcccc}
		\toprule
		\multirow{2}{*}{Dataset} & \multirow{2}{*}{Method} & \multicolumn{2}{c}{NN} & \multicolumn{2}{c}{SVM} \\
		\cmidrule{3-4}\cmidrule{5-6}
		& & FV & DF & FV & DF \\
		\midrule
		\multirow{5}{*}{UCF101}  & Real feature & 5.1  & 5.4  & 8.1  & 8.0 \\
		& cGAN-baseline							& 5.5  & 5.9  & 9.0  & 9.4 \\
		& cGAN+Inf 								& 13.8 & 15.2 & 14.9 & 18.7\\
		& cGAN+Co								& 20.6 & 23.0 & 21.2 & 25.1\\
		& cGAN+Inf+Co 							& \textbf{23.7} & \textbf{27.3} & \textbf{25.4} & \textbf{28.8} \\
		\midrule
		\multirow{5}{*}{CCV}  & Real feature 	& 11.9 & 11.6 & 14.1 & 13.8 \\
		& cGAN-baseline						 	& 12.5 & 13.9 & 14.5 & 15.1 \\
		& cGAN+Inf 							 	& 18.2 & 16.0 & 21.9 & 19.5 \\
		& cGAN+Co 								& 23.8 & 19.7 & 29.0 & 24.4 \\
		& cGAN+Inf+Co 							& \textbf{27.2} & \textbf{22.3} & \textbf{33.1} & \textbf{26.1} \\
		\bottomrule
	\end{tabular}
\vspace{-1.4em}
\end{table}
We conduct the ablation studies to further evaluate the effect of two components of our proposal and the results are exhibited in Table~\ref{tab:baselineTab}. For the sake of simplicity, we only report the results on UCF101 and CCV, since similar results are yield on the other two datasets. From Table~\ref{tab:baselineTab}, we can draw the following observations: (1) With the aid of semantic condition, synthetic video feature exceeds the real feature, suggesting semantic knowledge transfer is meaningful for ZSL problems. (2) Both semantic inference and MMICC can improve the performance of vanilla cGAN by a large margin, suggesting both intrinsic structure information of video feature and semantic correlation are vital for ZSL. (3) MMICC plays a major role in improving classification performance, revealing mutual information correlation is effective and robust for countering heterogeneity between visual and semantic domains. (4) Our proposal incorporates semantic inference and MMICC into an unified framework and achieves a significant improvement when compared to the performance of single component, demonstrating they complement each other for learning a robust visual-semantic alignment for zero-shot generalization.

Moreover, we investigate the performance of our model on alleviating hubness problem by qualitative illustration. We randomly sample 8 unseen categories from Olympic Sports and visualize both original and synthetic FV feature yield by different methods with t-SNE~\cite{maaten2008visualizing}.

As shown in Fig. \ref{fig:visualize}, (a) illustrates the distribution characteristics of original FV feature. In (b), vanilla cGAN conditioned on word-vectors is hard to synthesize FV feature with high discriminative power, owing to video feature of different classes mix together, resulting in a severe hubness problem. In (c), when cGAN is equipped with semantic inference, the synthetic feature starts to show clustering properties, but instances of different classes still overlap with each other. In (d), MMICC can improve discriminative power of the synthetic feature to a large extent, which proves the effectiveness of mutual information constraint. Finally, in (e), both semantic inference and MMICC are adopted to achieve a best performance for mitigating the hubness problem. Compared to original feature, synthetic feature yield by our framework even suffer less hubness problem intuitively.

\section{Conclusion}
In this paper, we have adopted deep generative model GAN to overcome the limitations of explicit projection function learning in ZSL video classification. The distributions of video feature and semantic knowledge are fully utilized to facilitate visual feature synthesis. We have proposed semantic inference and mutual information correlation to endow conventional GAN architecture with zero-shot generalization ability. Synthetic feature proved to own high discriminative power and suffer much less information degradation issue than previous methods. State-of-the-art zero-shot video classification performance is achieved on four video datasets.

\section{Acknowledgments}
This work was supported by National Natural Science Foundation of China under Grant 61771025 and Grant 61532005.
\bibliographystyle{named}
\bibliography{ijcai18}

\end{document}